\documentclass[journal]{IEEEtran}
\usepackage{rotating}
\usepackage{graphicx}
\usepackage[table,xcdraw]{xcolor}
\usepackage{tikz}
\usepackage{capt-of}
\usepackage{algorithm}
\usepackage[utf8]{inputenc}
\usepackage[noend]{algpseudocode}
\usepackage{epsfig,array}
\usepackage{amsmath,amssymb}
\usepackage{amsfonts}
\usepackage{graphicx}
\usepackage{color}
\usepackage{epstopdf}
\usepackage{mathtools}
\usepackage{float}
\usepackage{subfigure}
\usepackage{placeins}
\usepackage{balance}
\usepackage{hyperref}
\usepackage{accents}
\usepackage{adjustbox}
\usepackage[margin=2cm]{geometry}
\usepackage{rotating}
\usepackage{dblfloatfix}

\usepackage[english]{babel}

\usepackage[sorting=none, style=nature]{biblatex} 
\newtheorem{theorem}{Theorem}
\usepackage{xspace}

\usepackage{biblatex} 
\ifCLASSINFOpdf
\else
\fi
\begin{document}
%
\title{ON-DEMAND-FL: A Dynamic and Efficient Multi-Criteria Federated Learning Client Deployment Scheme}
\author{Mario Chahoud$^{1}$, Hani Sami$^{2}$, Azzam Mourad$^{1,4}$, Safa Otoum$^{3}$, Hadi Otrok$^{1,5}$, Jamal Bentahar$^{2}$, and Mohsen Guizani$^{6}$ \\
	\normalsize $^{1}$Cyber Security Systems and Applied AI Research Center, Department of CSM, Lebanese American University, Lebanon  \\
	\normalsize $^{2}$Concordia Institute for Information Systems Engineering, Concordia University, Montreal, Canada\\
	\normalsize $^{3}$College of Technological Innovation, Zayed University, Dubai, UAE\\
	\normalsize $^{4}$Division of Science, New York University, Abu Dhabi, UAE\\
	\normalsize $^{5}$Center of Cyber-Physical Systems (C2PS), Department of EECS, Khalifa University, Abu Dhabi, UAE\\
	\normalsize $^{6}$Mohammad Bin Zayed University of Artificial Intelligence, Abu Dhabi, UAE\\
}
\maketitle
\thispagestyle{plain}
\pagestyle{plain}
\begin{abstract}
In this paper, we increase the availability and integration of devices in the learning process to enhance the convergence of federated learning (FL) models. To address the issue of having all the data in one location, federated learning, which maintains the ability to learn over decentralized data sets, combines privacy and technology. Until the model converges, the server combines the updated weights obtained from each dataset over a number of rounds. The majority of the literature suggested client selection techniques to accelerate convergence and boost accuracy. However, none of the existing proposals have focused on the flexibility to deploy and select clients as needed, wherever and whenever that may be. Due to the extremely dynamic surroundings, some devices are actually not available to serve as clients in FL, which affects the availability of data for learning and the applicability of the existing solution for client selection. In this paper, we address the aforementioned limitations by introducing an On-Demand-FL, a client deployment approach for FL, offering more volume and heterogeneity of data in the learning process. We make use of the containerization technology such as Docker to build efficient environments using IoT and mobile devices serving as volunteers. Furthermore, Kubernetes is used for orchestration. The Genetic algorithm (GA) is used to solve the multi-objective optimization problem due to its evolutionary strategy. The performed experiments using the Mobile Data Challenge (MDC) dataset and the Localfed framework illustrate the relevance of the proposed approach and the efficiency of the on-the-fly deployment of clients whenever and wherever needed with less discarded rounds and more available data.
\end{abstract}

\begin{IEEEkeywords}
IoT, Federated Learning, Privacy, Client Selection, On-Demand Client deployment, Containers, Docker, Kubernetes, Kubeadm.
\end{IEEEkeywords}

%
\IEEEpeerreviewmaketitle

\section{Introduction}

\IEEEPARstart{A}{rtificial} Intelligence (AI) is the brain of smart systems which became the core of our daily lives through close interactions. With endless opportunities when using AI technologies, researchers look forward to develop robust and intelligent models to serve the community \cite{sarhadd}. Being the main pillar of smart technologies, data generated by smart devices should be collected from various entities and locations while maintaining top of the line privacy during data collection, transfer, training, and models deployment in production. Researchers try to maximize the quality of service (QoS), security \cite{touficwithdrazzam}, and communication quality in future smart cities \cite{mario20}. Today's users have access to mobile and edge devices \cite{haninewithdeazzam}, which provide vast amounts of data that can be used in numerous AI applications \cite{kadoumm}. However, privacy concerns are limiting the public's access to such data, particularly when it comes from individual smartphones or Internet of Things (IoT) devices, where researchers try to optimize multiple aspects related to fog and IoT devices using these data \cite{mauricekhab}. Additionally, the restriction on sharing private data has an impact on the process of developing and refining AI systems.

The recently used FL scheme \cite{mario12} helps mobile devices train machine learning models using their own data. The process includes selecting a group of accessible devices to take part in a learning phase, which is repeated numerous times until the model converges. The clients communicate the new weights to the server at the end of each round, where they are aggregated to create the overall model. The mechanism used by default to choose the clients who will take part in each round is randomness. However, selecting some clients could result in dropping out before finishing the learning rounds when working with devices that have restricted capabilities and resources. Additionally, a high volume of transferred updates may make communication between the clients and the server expensive and slow. Choosing a trustworthy group of customers in this situation can assist reduce redundant gradient information and improve network congestion \cite{mario4}. Additionally, the distribution of classes in federated learning is unbalanced, with some classes outnumbering others. Therefore, selecting clients with similar class labels may lead to biased and flawed models \cite{oamr2}.

The literature provides solutions for client selection that handle the majority of the aforementioned issues using a variety of approaches. However, none of the prior work proposes on-demand client deployment in the context of federated learning, by deploying clients instantly in extremely dynamic situations. Devices with poor learning resources and capabilities are missing out on the opportunity to participate in learning. However, new devices can easily and effectively participate in learning thanks to containerization technologies. Moreover, smart devices are busy at specific period of the day \cite{mario11}. The availability and resources needed in the area of interest can thus be provided anytime, anyplace, through the on-demand system. A tool for efficiently creating Kubernetes clusters is called Kubeadm. These clusters can be created using any resource-constrained device with the help of a master node.

Containerization technology is more lightweight \cite{mario13} than virtual machines. For that, the containerization technology has been used to deploy some services on fog devices near the IoT users \cite{mario1} for maximizing the quality of services. By suggesting an on-demand and on-the-fly client deployment method for federated learning based on Kubeadm and Docker, we take advantage of containerization technologies in this work. Due to the limited number of available clients, our suggested architecture enables a rapid deployment of new clients to participate in environments with data shortages. Moreover, our plan takes into account the device motions and locations, which have a significant impact on the quantity and heterogeneity of the data generated. Additionally, the on-the-fly deployment configures users for learning by taking into account their areas of interest. To illustrate the contributions and advantages of our approach, we use the MDC dataset \cite{mario14}, to represent real life simulation. The LocalFed \cite{marioo15} helps in applying FL models and to be integrated with the proposed genetic algorithm solution. We take real life scenarios that require on the fly client deployment to serve some machine learning models. First, we assume that there are a considerable number of devices moving (i.e., a lot of relevant data) in a client environment without any learning capabilities. Deploying containers on devices situated in these locations is quite advantageous in this regard to utilize such a vast volume of created data. Second, we assume that the environment is not entirely set up for learning and embedding volunteer client devices that are active in the neighborhood. In this situation, while taking into account our objective functions and limitations, our on-the-fly deployment technique enables proficient clients to engage effectively in the learning process and environment. Such scenarios motivate the work of \cite{hanialso} where more intelligent resource provisioning is required in such areas. The tests conducted in this paper produced encouraging outcomes in terms of fewer rounds, higher volume and heterogeneous data, improved accuracy in the early stages, less discarded rounds in the learning, and deployment of on-demand models and clients anywhere, anytime. The contributions of this paper are summarized as follows:
\begin{enumerate}
\item An innovative method of on-demand client deployment and selection that addresses the issue of client availability in pre-configured FL areas and places with a shortage of clients with the necessary capabilities and resources for learning.

\item An effective deployment and orchestration of services and models for machine learning on the IoT and Mobile devices of the recently created volunteering clients.
\item Formulating the deployment as a multi-objective optimization problem.
\item Implementing the Genetic Algorithm (GA) to gain from its evolutionary strategy while solving the deployment problem. \\

\end{enumerate}

The rest of this paper is organized as follows. In section II, we present the literature review. In section III, we present the architecture and methodology of the proposed client deployment approach. Section IV represents our On-Demand-Fl Formulation. In section V, we present the genetic algorithm. In section VI, we illustrate the experiments and the evaluation of obtained results followed by the conclusion in section VII.


\section{LITERATURE REVIEW}

Recently, researchers have made significant contributions to federated learning's communication, cost, security, privacy, and accuracy \cite{mario16}. In this section, we provide an overview of the recent related work in the literature.

The authors of \cite{review9} emphasized the significance of managing IoT resources while installing services on them utilizing lightweight technology like Docker containerization. A fog-formation on-demand architecture was given by the authors in \cite{mario1}. Fog devices often have specified locations and conduct specific services. The authors used containerization technology to deploy containers quickly and install the required services on the devices in order to increase device availability. Additionally, based on this design, fog devices can now be installed and made available to use whenever and wherever volunteers are located. The developers of \cite{mario2} extended on this architecture by proposing an on-demand micro-services deployment with the lowest possible cost while maintaining reachability between the desired customers and the accessible vehicular fog clusters. Furthermore, the authors in \cite{haniwithjamal1} tried to operate the placement of services proactively using Deep Reinforcement Learning.
We make use of this architecture in our work to dynamically allocate and deploy prepared clients to take part in a federated learning round using containers.

In \cite{mario4}, the authors find a solution to the issue of a bottleneck when there are a lot of customers in FL. The authors suggested a system that only chooses a few clients that are carrying representative gradient data to send changes to the server. Submodularity was used by the writers of this research to encourage the diversity of the chosen clients, which helped them achieve their goals. According to \cite{mario5}, it is not necessary to select a maximum number of clients throughout each round as most of the proposed client selection approaches concentrate when selecting the maximum number of clients who can update their models. We can reduce training loss and increase accuracy by choosing fewer clients in the early rounds and more in the latter ones. The authors of \cite{mario6} suggested a dynamic optimization based on a trade-off between increasing selection and lowering participant clients' energy usage.

Using optimization techniques, some strategies \cite{mario7} try to balance the trade-off between the client consumption and the learning rate. The authors in  \cite{mario9} developed a new sampling technique that combines customers into pools for easy selection and summarizes their attributes in a single floating point. In addition, the suggested model achieves 50–80\% faster convergence in a high-class imbalance and a low-data environment.
Authors in \cite{review1} attempted to improve the routing protocols' level of service and privacy. They were successful in segmenting the routing domain into advantageous subdomains, and federated reinforcement learning is used to maintain privacy. The authors in \cite{review2} proposed an effective FL-based algorithm for anomaly detection. Multiple local deep reinforcement learning models were used to build the FL, which prevents privacy leakage while also suggesting a leaking technique to boost prediction. Additionally, authors of \cite{review3} utilized FL and Blockchain to offer a trustworthy and secure energy solution for trade between numerous businesses. The authors in \cite{review4} suggested deep federated learning for health care data. Recently, researchers started to focus on healthcare systems \cite{alaa}. For instance, in order to diagnose COVID-19, \cite{review6} suggested a clustered FL method.

For selecting clients, the authors of \cite{mario10} suggested a multi-criteria model. The system takes into account the clients' resources to forecast the amount of time needed to complete the jobs without stopping the process until convergence. Additionally, according to a report by \cite{mario11}, IoT and mobile devices can get quite active at certain times of the day. As a result, the number of clients participating in the FL process significantly relies on the time of day, with a low number during the day and an increase of 4x at night that may result in the discarding of certain learning rounds. In order to guarantee a balanced, impartial model, the developers in \cite{review5} suggested a new scheduling architecture based on the historical participation of each edge device.

The authors of \cite{mario4}, \cite{mario7}, and \cite{mario10} considered resource capacities. \cite{mario11}, \cite{mario5}, and \cite{mario6}  all focused on the trade-off between the amount of clients chosen in each round, while \cite{mario11} and \cite{mario10} also considered the time of day for choosing clients, and communication costs were taken into account by the authors in \cite{mario10}. Authors in \cite{review1}, \cite{review2}, \cite{review3}, and \cite{review4} target the use of FL in applications without taking into consideration the availability of enough clients.

If there are not enough clients available to participate in FL, none of the methods for selecting and optimizing client selection described previously is applicable. In this paper, we focus on the issue of client availability in various areas. To the best of our knowledge, no prior research work has focused on the ability to execute machine learning models in environments that are not set up for training or do not have the flexibility to add new devices. Furthermore, none has provided the ability to deploy clients instantly whenever and wherever there are volunteer devices. This paper concentrates on areas with lots of data that helps improve the model, while making more clients available for learning.

\section{ARCHITECTURE AND METHODOLOGY}
In this section, we present our on-demand FL architecture, followed by descriptions of the functionalities in each component of the proposed architecture.
\subsection{Architecture Overview}

Mobile and IoT devices provide a lot of data in a highly dynamic environment, including photos, location histories, the whereabouts of autonomous vehicles, and captured images. Full access to such data would allow for the development of powerful machine learning models that could provide reliable and intelligent services. However, security and privacy are major concerns for data owners. FL has been suggested in this context as a decentralized solution that combines intelligence with privacy. It comprises of a machine learning model that is shared by a number of clients and uses locally learned data. Initially, the server generates a model to serve a certain task and then asks a random set of clients of size $K \times C$ to participate in each round and exchange parameters with the server, where $K$ is the number of available clients, and $C$ is the portion of clients considered. The chosen model is then trained on each selected client's local data, and the new derived parameters are sent to the server. The server then combines the gathered parameters and calculates a FedAvg to build upon and improve the overall model. Multiple rounds of this technique are carried out to improve convergence and precision. However, failing from or report later the new parameters causes the server to end the round, which could slow down or stop the learning process altogether. Therefore, the chosen clients must be available throughout a FL training round. The quantity of resources that are accessible throughout the entire round serves as the definition of this availability. Additionally, a learning environment must be set up by connecting to the server, gathering the necessary materials, installing them, and beginning the training procedure.  

\begin{figure}[]
	\centering
	\includegraphics[width=0.5\textwidth]{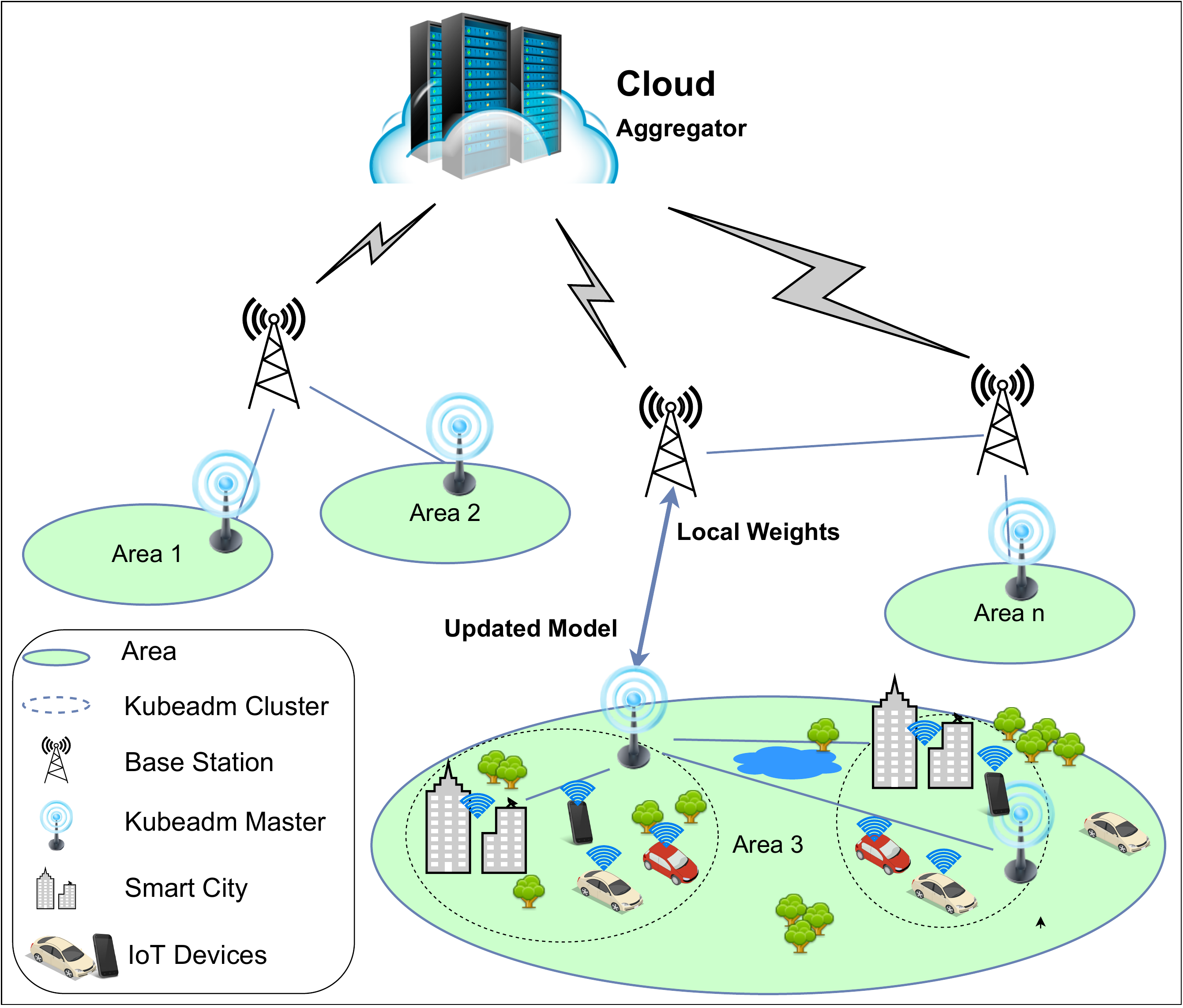}
	\caption{Overall Architecture.}
	\label{fig}
\end{figure}

Figure \ref{fig} provides an illustration of the proposed framework's architecture. From a collection of constrained devices, Kubernetes clusters may be effectively built using the Kubeadm tool. In this work, we suggest creating the cluster on-the-fly using the server and the devices that are willing to volunteer. The server or service provider, orchestrators or small servers, and user devices make up the three levels that constitute the architecture. In the following section, we go over how each layer fits into the overall architecture flow.

According to our architecture, the server is in charge of providing clients with container images, managing the global model, and keeping a safe connection with the lower levels. Additionally, with the help of the orchestrator, the server is in charge of making decisions as they arise regarding the participation of nodes in cluster formation. The Kubeadm clusters must be created by the orchestrators, who are also in charge of managing container deployment, tracking device movements in their area, and adding new clients to the cluster. In cases of high mobility, the orchestrators are also in charge of handling client deployment and selection as well as sending requests to the server for container deployment on clients. The orchestration layer's key benefits are its ability to reduce the risk of a single point of failure and to speed up operations, especially in highly dynamic circumstances. Additionally, having an orchestrator for each client group manages server resources and helps cut down on network connection overhead.
For FL, when data is used without sharing, the volunteering fog devices act as clients. With the help of lightweight containers and our on-demand architecture, clients may execute a machine learning model whenever and wherever they want. The related service will not be running on any device besides those taking part in the FL rounds.

Each node in our design has its components built and pushed to the Docker Hub repository to create images that can be installed on participating clients. The Kubeadm Containerization Required Modules must run on all the nodes in our architecture (i.e., Server, Orchestrator and Fog nodes). On the server, the Aggregator, Client Deployment, Kubeadm Environment Initializer, Oracle Engine, Communication Manager, and Orchestrator Manager are all in use. Both the orchestrator and the client devices run the Client Profiler. The orchestrator is used to operate the Client Manager and the Client Deployment components. Finally, the Fog Client and the Learning Respond components run on the client nodes.

\begin{figure}[]
	\centering
	\includegraphics[width=0.5\textwidth]{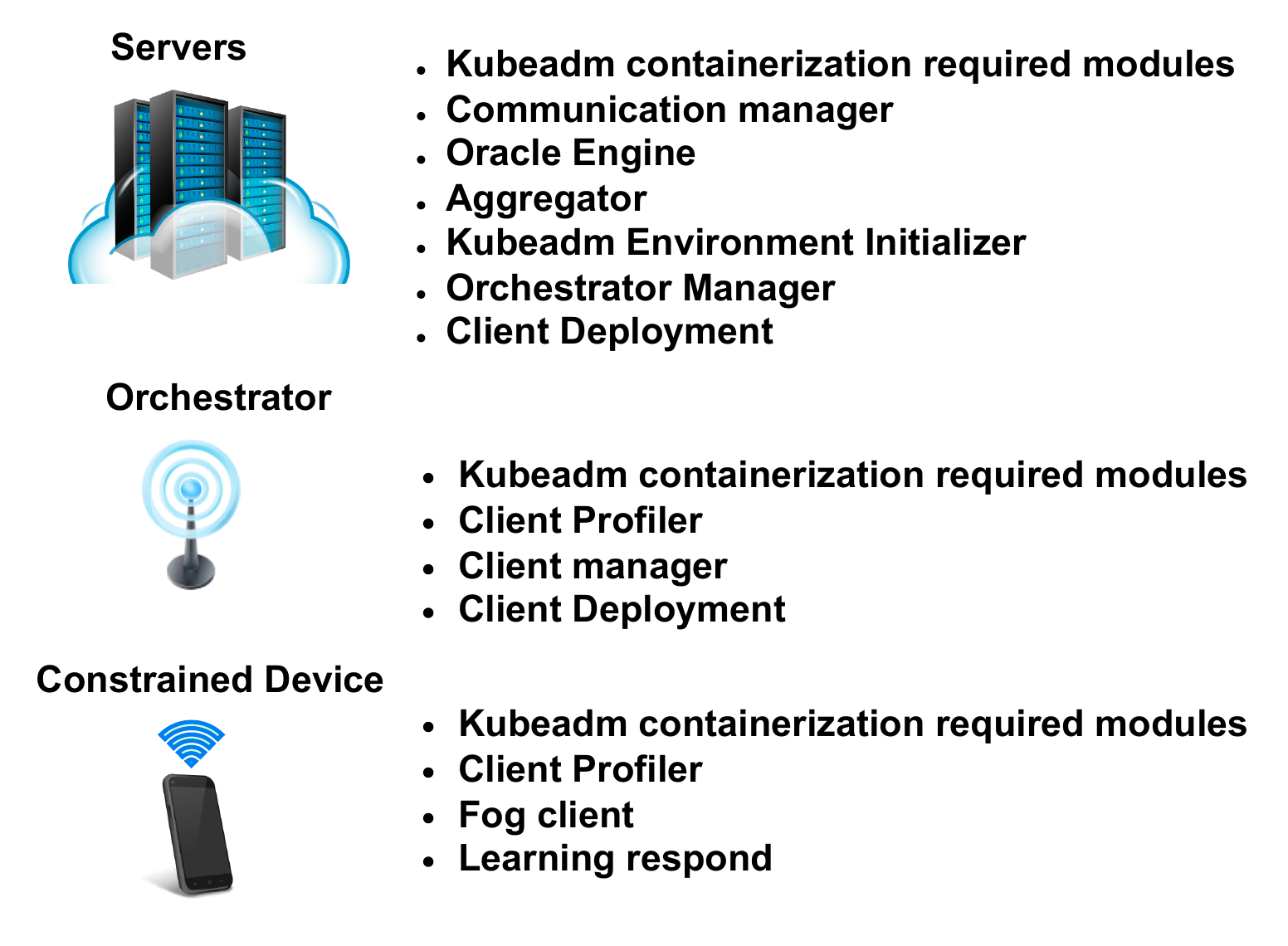}
	\caption{Overall Node Architecture Per Layer.}
	\label{figa}
\end{figure}

The below subsection describes the functionalities of each sub-component represented in Figure \ref{figa}
\subsection{Communication Manager}
The communication between the server, orchestrator, and deployed clients is handled by this component. In our architecture, the server sends a request to the deployed clients to establish a connection in order to exchange messages, broadcast the revised model weights, and exchange updated parameters. The handling of communication is a critical aspect to take into consideration in FL \cite{communicationandprotocol}.
\subsection{Oracle Engine}
This component is responsible for selecting and building the best machine learning model that fits the environment \cite{bestmlchoose}, which is later sent to the clients. 
\subsection{Aggregator}
In order to optimize the federated aggregation function, the aggregator must aggregate the newly received weights in order to update the global model. Additionally, the aggregator verifies the number of updated weights received, as constrained devices may malfunction or indicate delays. As a result, this component takes that into account and discards the round if the number of updated weights received falls below a certain limit set by the server. Moreover, different aggregator functions yield to different results \cite{omarr}. This component is responsible to detect and select the best one based on the previous experience in each area.

\subsection{Client Profiler}
With the help of this module, the orchestrator is able to get an overview of all the devices that are willing to offer their services by gathering data on their battery life, disk size, memory, CPU usage, location history, and average time spent in a given area. This module updates the data about resources and volunteers. Using the updated list, better cluster formation and placement choices are made.

\subsection{Kubeadm Containerization Required Modules}
The Kubeadm tool aids in the quick and secure construction of Kubernetes clusters. Docker, which must always be running on all devices, is supported by Kubeadm. Kubeadm's capacity to function on a variety of machines is a plus. The command line known as Kubectl must be deployed in order to manage communications, keep clusters alive, and verify the condition of devices and services. Images are fetched, initialized, and started after the client devices and the orchestrator/master node have been set up.

\subsection{Kubeadm Environment Initializer}
On the mini-servers that are accessible in each area, the server first starts by initializing a few orchestrators. The first orchestrators are chosen based on client device motions that the server has only just begun to collect. The orchestrators are then in charge of carrying out this task. In an area where there are many devices that are willing to volunteer, the server first tries to push orchestrators there. Additionally, the server determines if the day is a weekend or a weekday so that it can determine earlier on where to place some orchestrators. In the context of this article, we assume that a list of potential mini-server orchestrators is accessible. In summary, the server (1) chooses the locations where Kubernetes clusters will be created and some of its devices will receive container deployments, and (2) when the orchestrator node is initially configured, adds some willing devices to the cluster. All of the devices that can volunteer in the region where the orchestrator is created are prompted by the server to join the Kubeadm cluster. The Kubeadm environment is now prepared to push services as needed without any setup lag. Additionally, the Master node must always be active alongside the volunteers. The cluster collapses if the master node is down. From this point forward, in the event that the first orchestrator goes down, the orchestrator or master node must always select a different available mini-server in the area to switch to. As a result, there is no longer a need to start again with a fresh Kubeadm cluster and waste time initializing everything. We presume that each region has an enough number of mini-servers. If no mini-servers are accessible, the module may select a client device to act as an orchestrator for a brief period of time while taking into account its profile information. This process should be written as a multi-objective optimization problem, which is outside the scope of this study but will be taken into account in subsequent work. For now, we consider that the chosen nodes to serve as orchestrators are suitable to run the tasks.

\subsection{Client Manager}
The functions of the client manager operating on the orchestrator nodes are covered in this section. The orchestrator must first keep an eye on its clients' movements. An orchestrator will ask the server to deploy containers and some of the chosen clients in its area if it sees significant movement in the area. High movements generate a large amount of data, which helps a model be better trained. The client manager also keeps track of how many rounds each user has served. Our technique prevents starvation in client selection for subsequent rounds by keeping track of how many rounds a client has participated in overall. Additionally, this module handles a node's remaining time in the cluster and examines its average staying time in a region to help us reach our training goal. When making the selection in the following round, it's crucial to take the cluster's average remaining time into account. Additionally, the client manager notifies the server of the availability of any new client devices that enter the area.
\subsection{Orchestrator Manager}
The responsibilities of the orchestrator manager operating on the server are covered in this section. It first records the clients who have chosen and used orchestrators. These statistics show how users behaved in a specific location. Additionally, selecting the node as an orchestrator raises some security issues, but this subject is outside the purview of this work. We assume that the server has confidence in the accessible mini-servers. To aid in the selection of the orchestrator, the orchestrator manager uses the historical data of users who have already trained some models \cite{mario2}. The movements of the clients should be taken into account when choosing the orchestrators. The server also considers whether it is a weekend or a weekday, and on weekends it deploys some orchestrators in areas with certain mountainous terrain where people may be more active. The server performs this at regular intervals, and each orchestrator tracks its clients' movements after that. In order for the server to determine the areas of interest, this module also handles requests from orchestrators to deploy additional containers in their area. Additionally, this module is in charge of choosing clients from various areas in order to guarantee data heterogeneity and manage class imbalance.
\begin{figure}[]
	\centering
	\includegraphics[width=0.5\textwidth]{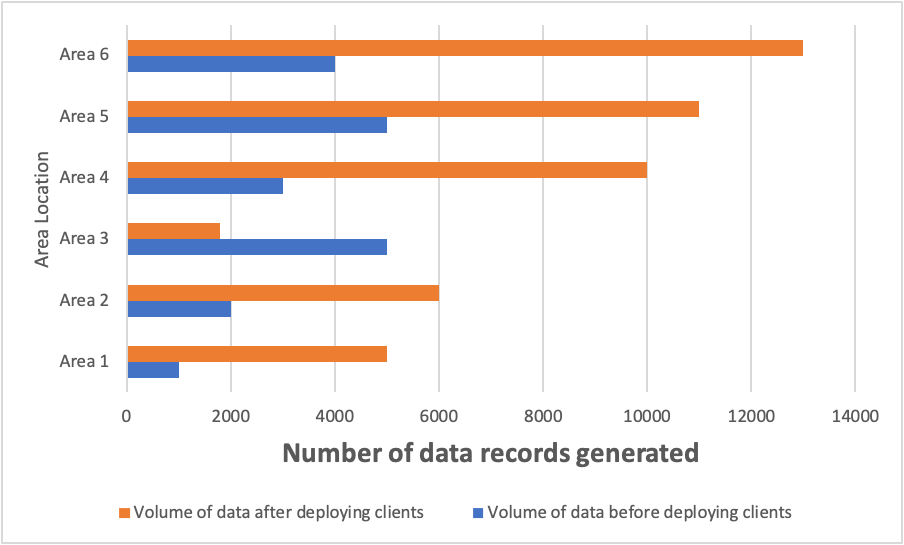}
	\caption{Average Data Volume during the process of learning before and after deploying on-demand new clients.}
	\label{fig4}
\end{figure}
\subsection{Fog Client}
The volunteer devices are running this module. Sending a request to the orchestrator with its profile information is one of its functions. Additionally, this component keeps track of the machine learning algorithm that is operating on each end and notifies of any changes to the orchestrators.

\subsection{Learning respond}
This module is responsible of preparing the data set of the learning service as data fetching, data pre-processing, and splitting the data into train and test.

\subsection{Client Deployment - Decision Module for Selecting and Deploying Volunteering Clients} 
This module is triggered by the orchestrators to efficiently deploy clients in an area. The efficient deployment of clients is applied while taking into consideration different criteria, which is considered multi-objective problem. This problem is presented and implemented in sections IV and V.

\section{CLIENT DEPLOYMENT PROBLEM}

In the first part of this section, we define our multi-objective optimization problem for client selection and models' deployment. Followed by mathematical representations of the input, output, constraints, and objective functions.

\subsection{Problem Definition}
Given a set of available volunteering devices $C_1, C_2, \dots, C_n$ along with their utilities for learning $U_1, U_2, \dots, U_n$, we have to find the best portion of clients to deploy in each learning round  while taking into consideration different aspects in terms of resource consumption, priority, area location, availability, and movements. Selecting the best set of clients and optimally deploy them for learning is complex.

In the single Knapsack problem \cite{knapsack}, items with different weights and values are given, in addition to the Knapsack capacity. The main objective is to pick items that maximise the profit formula while taking into consideration weight limits. In multi-objective Knapsack optimization, there are $p$ objective functions simultaneously, where the objective is to maximize the profit in all cases subject to different constraints.

Given $b$ items with characteristics like: weights, volume, and profit, we need to select some items that maximise the total profit $p$ without exceeding the knapsack capacities. 

The mathematical representation of a multi-objective knapsack with numerous constraints is shown in Eqn (\ref{eqn:k1}).
\begin{equation}
\label{eqn:k1}
   max(z_k(x)) = \sum_{i=1}^{b} c_{k}^{i} x_{i}
\end{equation}\
$x \in X$ where $X$ is the set of feasible solutions, and $z_k(x)$ represents the $k^{th}$ objective function. $c_{k}^{i}$ represents a profit of item $i$ under the $k^{th}$ objective function.

subject to:
\begin{equation}
    \sum_{i=1}^{b} w_{j}^{i} x_{j} \le W_{i}
\end{equation}

Where $x_{i}$ is = 1 if $X_{i}$ is picked in the solution. $W$ represents the overall knapsack capacity and $w_{j}^{i}$ represents a weight / cost of item $i$.
\begin{theorem}
Our multi-objective optimization problem is NP-Hard
\end{theorem}

The multi-objective optimization problem described in section V can be proved to be NP-Hard by a reduction to the famous Knapsack Problem, specifically to the multi-objective knapsack problem as shown below:

\textbf {Proof}: Given an instance of a client deployment problem, we reduce it to knapsack as follow: \\
Clients are the items of a knapsack, the features of a client such as priority, movements, availability, and resources capacity, are passed to the objective functions to represent the fitness of a client (value of an item). The weights of an item are represented as the cost of selection, such that the client must have enough time to finish a round $C_{availability_i} \ge T$, certain movement patterns, and resources consumption. In conclusion, the objective is to maximize the fitness (profit) of selected clients subject to constraints listed later in this section.

This reduction yields that our client deployment problem is NP-Hard.

\subsection{Problem Formulation}
The objective of our problem is to optimize the number of active deployed clients, diversity of the data, the volume of the data, quality of learning, and serving on-demand requests.

    \subsubsection{Input Data}
    \hfill \\ 
    In our problem formulation, we have a set $C$ of available host devices in different areas. Containers should be deployed on a portion of these hosts to be included in a learning round.
    The set of available hosts is represented as a matrix $ C \in \mathbb{R} ^ {n \times 6} $ corresponding to six input features. Each host is offering its resources as CPU, memory, disk space, battery life, availability, and area location as follows: \\
    \begin{center}
        
    $C_{i} = [C_{CPU_i}, C_{memory_i} ,  C_{diskspace_i} ,C_{battery_i} ,C_{availability_i} ,C_{area_i}]$ $\forall i \in \{1, \dots, n\}$ where :
    
    $C_{CPU_i}$ : CPU availability on $C_{i}$.
    
    $C_{memory_i}$ : Memory availability on $C_{i}$.
    
    $C_{diskspace_i}$ : Disks pace availability on $C_{i}$.
    
    $C_{battery_i}$ : Battery level on $C_{i}$.
    
    $C_{availability_i}$ : Availability time of $C_{i}$ in a specific area location.
    
    $C_{area_i}$ : Current area location of $C_{i}$.
    
\end{center}
\begin{figure}[]
	\centering
	\includegraphics[width=0.5\textwidth]{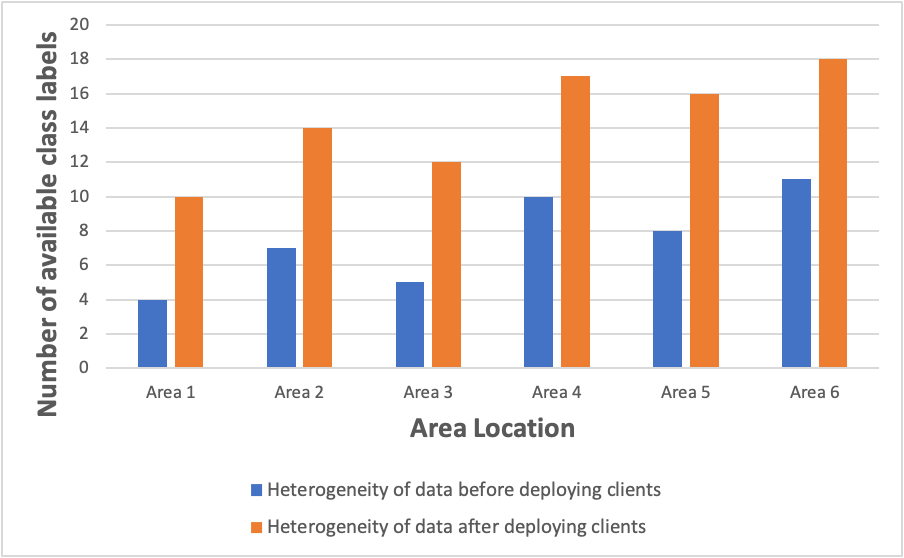}
	\caption{Average number of class labels during the learning process before and after deploying new on-demand clients.}
	\label{fig5}
\end{figure}
    Each client has a utility input from matrix $U \in \mathbb{R} ^ {n \times 4}$, which corresponds to four input features. Each utilization $U_{i}$ represents the CPU, memory, battery, and disk space consumption of the service on hosts extracted from the google cluster workload dataset \cite{Googlecluster} for a client $C_{i}$
    \begin{center}

    $U_{i} = [U_{CPU_i}, U_{Memory_i}, U_{Battery_i}, U_{diskspace_i}] $

    $U_{CPU_i}$ : CPU consumption of client $C_{i}$.
    
    $U_{Memory_i}$ : Memory consumption of client $C_{i}$.
    
    $U_{Battery_i}$ : Battery consumption of client $C_{i}$.
    
    $U_{diskspace_i}$ : Disk space consumption of client $C_{i}$.\\
      \end{center}
      
    The orchestrators are responsible to monitor and report the movements of clients represented in an array of Movements $Movements_{i} \in \mathbb{R}$ illustrating the rate of visiting areas of each client in addition to an array $A_{k}\in \mathbb{R}$ $\forall k \in \{1, \dots, m\}$, where $m$ is the number of area locations. $A_{k}$ represents if orchestrators on-demand request to deploy clients in specific areas by having a value of 1 if requested and 0 otherwise. 
    \\
     \subsubsection{Output Data}
    \hfill \\ 
    A set of chosen clients participating in a learning round represented in $K$ as a one-dimensional array. $K_{j}$ represents a client device in the output array, where $j$ is the device number, $\forall j \in \{1, \dots, n\}$ $K_{j}$ $\in$ $0,1$. If $K_{j}$ is chosen to be deployed in a learning round will have a value of 1 otherwise, it is 0.
    \\
     \subsubsection{Constraints}
     \hfill \\ 
     \textbf{Resource constraints:} 
    A client $C_{i}$ is selected if its resource capabilities do not reach their full capacity after deployment. The hosting capacity is represented as: CPU, memory, disk space, and battery level along with the learning utilization consumption in terms of CPU, memory, disk space, and battery consumption. When the FL service is deployed on a client, each client $C_{i}$ has a utilization $U_{i}$ that represents how much that service uses from the resources of that client. Below is the mathematical representation of this constraint: 
    
    \begin{equation}
    U_{CPU_i} \times K_{i} \le C_{CPU_i}
    \end{equation}
    \begin{equation}
    U_{memory_i} \times K_{i} \le C_{memory_i} 
    \end{equation}
    \begin{equation}
    U_{diskspace_i} \times K_{i} \le C_{diskspace_i}
    \end{equation}
    \begin{equation}
    U_{Battery_i} \times K_{i} \le C_{Battery_i}
    \end{equation}
    
    $\forall i \in \{1, \dots, n\}$ i.e. for all available clients $C_{i}$ and their utilization $U_{i}$.\\
    
   \textbf{Minimum availability time:}
    This is to avoid devices dropping from a learning round due to their movements and staying period in an area/cluster. For that, the server decides on a parameter ${T\in N}$ that represents the minimum time needed for one round. Therefore, a client can be deployed and selected if $C_{availability_i}$, which is the staying time of $C_{i}$ in its area, is greater than $T$.
    
    \begin{equation}
    \forall i \in \{1, \dots, n\} \;\;\;\;\; C_{availability_i} \ge T
    \end{equation}
    
   \textbf{Minimize client starvation:}
    One of the features that our model uses to optimize the deployment is client movements. This feature is monitored by the orchestrators and recorded as a counter for each client visiting a specific place. Based on this, choosing always high movements clients leads to similar selection every round. Henceforth, to avoid repetition in selection, the server chooses a threshold to determine the percentage of clients with high movements. A client is in high movement if $Movements_{C_i}$ is greater than $MaxT$, where $MaxT$ is the threshold to determine if a client is in high movement.
    
    \begin{equation}
        \sum_{i=1}^{n} K_{i} \le Mt \; 
        \forall i \in 1, .. , n, \; Movements_{C_i} \ge MaxT
    \end{equation}
    Where $Mt$ is the portion of clients to be selected with high movements.\\
    
     \textbf{Applying priorities to clients:}
    Each client has a priority number from 1 to $t$, where $t$ is a value between 1 and 10. Our model prioritizes the clients based on their contribution to learning by giving them more priority based on their local accuracy if not deployed. A higher priority means that this client must be deployed first. \\
    
    \textbf{Weights summation:} to provide more flexibility in prioritizing the objective functions, the method of adjustable weights \cite{hihi} helps by multiplying each objective by a decimal value between 0 and 1 where their sum (\ref{eq:1}) is equal to 1.
    \begin{equation}\label{eq:1}
        W_{f_1} + W_{f_2}+ W_{f_3}+ W_{f_4}+ W_{f_5} = 1
    \end{equation}

    \subsubsection{Objective Functions}
    \hfill \\ 
    
    In a federated learning (FL) application, the global objectives are to maximize the number of clients that can complete FL rounds without system crash while minimize the number of rounds. In our proposed architecture, we present the objectives of the selection and placement solution in the FL context as follows:
    \begin{enumerate}
    
   \item  Minimize the number of deployed clients:
    The objective is to minimize the number of deployed clients in the learning phase. 
    \begin{equation}
        F_1 = min((\sum_{i=1}^{n} K_{i}) \times W_{f_1} )
    \end{equation}
    
    Where $W_{f_1}$ is the weight of this objective function.
    
    Minimizing the number of active hosts saves energy and resource consumption, which means less battery, CPU, and memory consumption of the overall available clients. This results in increasing the availability of fog devices, and thus, other services or applications can use the extra amount of available hosts in an area. Moreover, in FL, the server and client devices struggle with the high rate of exchanging parameters and updating weights. Following this objective, the high congestion on the network is decreased. As already discussed in \cite{mario5}, it is unnecessary to select a maximum number of clients to participate in each round. Selecting fewer clients at the early stages and more clients in later rounds helps improving the training loss while reaching higher accuracy.\\
    
   \item  Maximize the volume of data:
    clients frequently visiting places in a specific period of time, generate more data than others having fewer movements or records of visiting places. Therefore, the objective is to maximize the deployment of clients generating high volumes of data. This objective is expressed as follows:
    \begin{equation}
        F_2 = max( (\sum_{i=1}^{n} K_{i} \times Movements_{C_i} )  \times W_{f_2})
    \end{equation}
    Where $W_{f_2}$ is the weight of this objective function. 
    
    In any machine learning algorithm, the more data there is, the better the model performs. Choosing clients with high movement in an area results in generating a large volume of data to be used, thus a better performance is achieved. Clients with high movements indicate a high frequency of visiting places monitored by the orchestrators in each area.\\
    
   \item  Maximize the quality of learning:
    We aim to maximize the accuracy of the model at early stages by choosing high-value clients based on their contribution to the model.
    \begin{equation}
        F_3 = max ((\sum_{i=1}^{n} K_{i} \times C_{P_i} )\times W_{f_3} )
    \end{equation}
    Where $W_{f_3}$ is the weight of this objective function and $C_{p}$ is the priority factor of each client.
    
    Checking the local accuracy and having a history of score of the un-deployed clients helps in extracting useful information about the kind of data these clients have. A higher accuracy means that the client's testing data is somehow good for the model. Therefore, including their training data will boost the model accuracy by deploying such clients at early stages. Maximizing the accuracy of a model results in better performance and quality decisions. In addition, having the accuracy high at early stages (rounds) helps real-time systems that are using the global model. \\
    
    \begin{figure}[]
	\centering
	\includegraphics[width=0.5\textwidth]{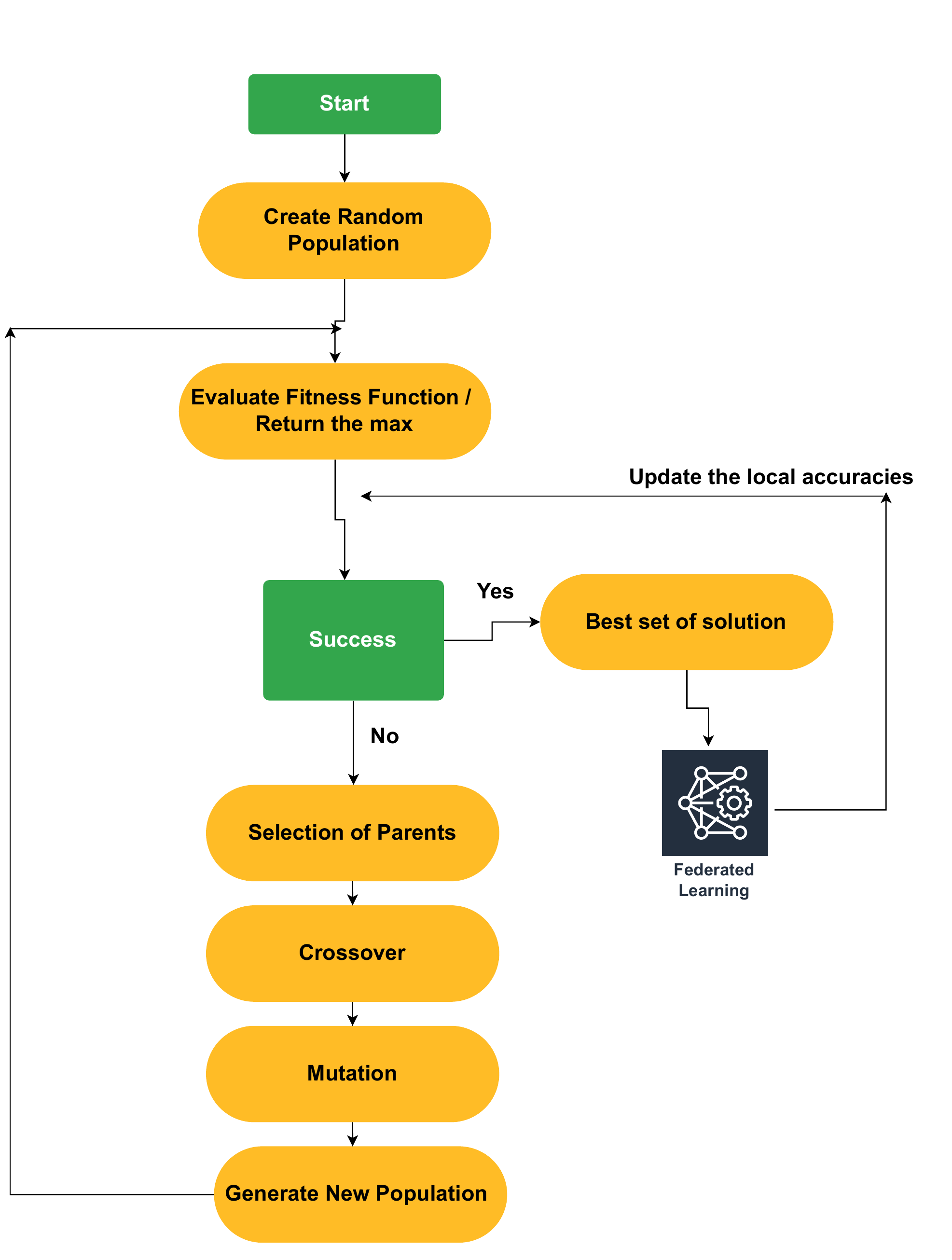}
	\caption{GA flow of activities}
	\label{fig10}
\end{figure}
   \item  Maximize the diversity of data:
    Federated learning suffers from Non-IID data, Therefore, the objective of this function is to maximize the diversity of data to limit and minimize the effect of Non-IID clients.
    
    \begin{equation}
        F_4 = max( (\sum_{i=1}^{n} K_{i} \times R_{i} ) \times W_{f_4} )
    \end{equation}
    Where $W_{f_4}$ is the weight of this objective function, and $R$ is the difference between the rates of the selected clients.
    
    A high score of $R$ indicates that selected clients are from different areas. Data generated from such clients is different and especially the output class features.
    A diversity in the output class in each round also helps the model to perform better on new testing data.\\
    
  \item  Maximize the serving requests:
    The objective is to maximize serving the requests sent from the orchestrators to deploy containers in some areas.

    \begin{equation}
    F_5 = max( (\sum_{i=1}^{n} K_{i} \times RT ) \times W_{f_5})    
    \end{equation}
    
    $W_{f_5}$ is the weight of this objective function and $RT$ is the average rate of clients selected from the requested areas calculated from the input array $A$ and the selection output $K$.
    
    The mini-servers / orchestrators monitor the average movements and locations of the users. When having high movements of clients in an area, the mini-server sends a request to the server to deploy containers due to high movement activities. The model should deploy containers on clients located in the requested areas. 
    \end{enumerate}
    
    After having the optimization functions declared, the multi-objective optimization problem becomes:
    
    $$Y = f(x) = [f_1(x), f_2(x), f_3(x), f_4(x), f_5(x)]$$
    
    Where:
    \begin{center}

    $F_1(X)$ = Number of active clients

	$F_2(X)$ = Diversity of the data.

	$F_3(X)$ = Mobility in an area.

	$F_4(X)$ = Quality of learning.

	$F_5(X)$ = Number of serving requests
    \end{center}
subject to:
\begin{center}

    $e_1(x)$ : Processing resources.

	$e_2(x)$ : CPU, memory, disk and battery resources.

	$e_3(x)$ : Minimum availability time.

	$e_4(x)$ : Avoid repetition and starvation of clients.

	$e_5(x)$ : Clients with high priority first.

	$e_6(x)$ : Sum of all objective function weights = 1.
\end{center}

\section{GENETIC ALGORITHM FOR ON-DEMAND-FL}
\begin{figure}[]
	\centering
	\includegraphics[width=0.5\textwidth]{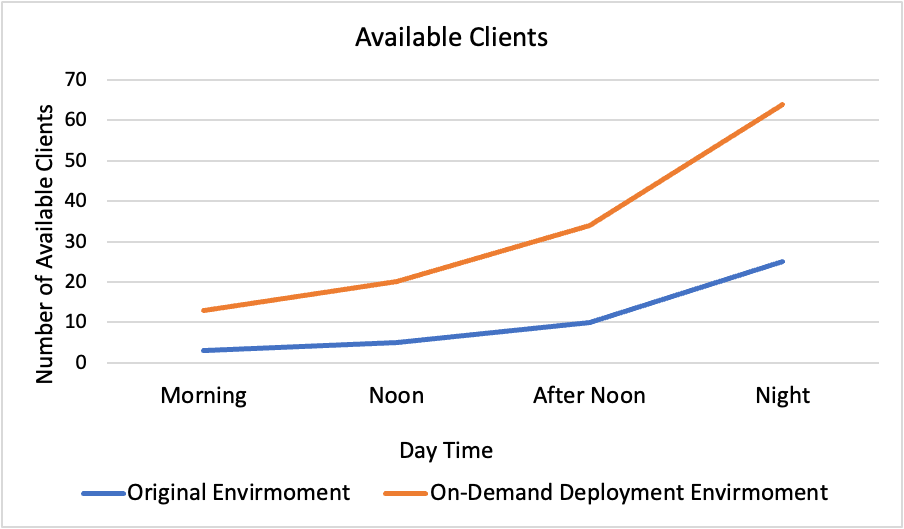}
	\caption{Average number of available Clients}
	\label{nbofclients}
\end{figure}
Multi-objective optimization problems are known to have multiple solutions and not only one optimal solution, these solutions are called $Pareto$ solutions. It is very important to obtain the $Pareto$ set solution in a short period of time. For that, using a Genetic algorithm (GA )\cite{GA} that is characterized by its evolutionary strategy is a good choice for such problems \cite{halaa}. GA imitates the natural selection process by choosing the fittest set of solutions for the reproduction of the next generations. Each chromosome in our GA solution is represented in a $K$ matrix. Each chromosome represented as $K_{j}$ illustrates the decision taken by the optimization model if client $K_{j}$ is deployed for a service by having a value 1, and 0 otherwise. $K_{j} \in$ [0 , 1].
\begin{algorithm}
\caption{Multi-objective genetic algorithm}\label{alg:cap}
 \textbf{Data:} Set of available clients
 
 \textbf{Result:} Pareto set approximation $P_{known}$
\begin{enumerate}
    \item Check if the problem has a solution
    \item Initialize set of solutions $P_{0}$
    \item $P'_{0}$ = repair infeasible solutions of $P_{0}$
    \item Update set of non-dominated solutions $P_{known}$ from $P'_{0}$
    \item $X \gets 0$
    \item $P_{X} \gets P'_{0}$
    \end{enumerate}
\begin{algorithmic}
    \While{(stopping criterion is not met),}
    \begin{enumerate}
    \setcounter{enumi}{6}
         \item $M_{X}$ = selection of solutions from $P_{X}\cup P_{known}$
         \item $M'_{X}$ = crossover and mutation of solutions of $M_{X}$ 
         \item $M''_{X}$ = repair infeasible solutions of $M'_{X}$ 
         \item increment X
         \item Update set of solutions $P_{known}$ from $M''_{X}$
         \item $P_{X}$ = fitness selection from $P_{X}\cup M''_{X}$
    \end{enumerate}
    \EndWhile

    \State\textbf{Return:} Pareto set approximation $P_{known}$

\end{algorithmic}
\end{algorithm}

The GA is presented in Algorithm 1. Our solution checks if containers can be deployed on available client devices. Next, a random sample of solutions $P_0$ is initialized by selecting clients to deploy. Afterwards, the obtained solution is evaluated and repaired for any violations of the constraints. The violations are repaired as shown in Algorithm 2. Reparation is performed by moving containers to other client devices if the machine learning exceeds
(1) the resource capacity of a client
(2) the client’s staying period in a particular area
(3) the "high movement" threshold decided by the server.
Non-dominated ($Pareto$) solution is generated in Step 4 of the (GA). Afterward, normal selection, crossover,
and mutation operators are applied, infeasible solutions are repaired, and finally, the
Pareto set is updated if improvements are possible. This process is repeated in multiple iterations. In our GA algorithm, we use binary tournament selection in order to choose individuals from populations. The crossover method used is the one-point crossover where the crossover is probabilistically applied for each parent, controlled by a hyper-parameter. Moreover, for the mutation phase, the bit string mutation is used. The approach gives every gene a mutation probability of $1/L$, where L is the number of clients for deployment. This approach protects the algorithm from falling in the local optimum, maximizes the diversity and it gives a fair opportunity for clients with a low probability rate to engage with the mutation. The flow of activities can be seen in Figure \ref{fig10}.

The complexity time of this algorithm can be divided into multiple parts. First, we have the total number of populations, then initialize the base population, fix constraints, calculate the fitness, selection of parent, cross-over, and mutation. Let $G$, $M$, and $N$ be the number of generations, chromosomes, and the number of nodes respectively. In addition to that,  let $C_{selection} = O(N)$, $C_{crossover} = O(N\times M)$ and $C_{mutation}=O(N\times M)$ be the complexity time of selection, cross-over and mutation process respectively. Moreover, Let $C_{fitness}$ and  $C_{fix-violation}$ be the complexity time of calculating the cost (fitness value) and to fix if any violations to our constraints respectively.
Therefore, the overall complexity time will be:
O(G$\times$($C_{selection}$ + $C_{crossover}$ +$C_{mutation}$ +$C_{fitness}$ + $C_{fix-violation}$)).

\begin{algorithm}
\caption{Infeasible solution reparation}\label{alg:cap1}
 \textbf{Data:} Infeasible Solution $K$
 
 \textbf{Result:} Feasible Solution
\begin{algorithmic}
\State feasible = False; i = 0
\While{$i \le n$ and feasible == False} 
\If{$K_{i}$ is overloaded} \Comment{check if overloaded}
\State deploy the container on another capable client
\State in the same area location
\EndIf

\If{$K_{i}.availability \le T$} \Comment{check staying time}
\State deploy the container on another not
\State chosen client with high movements record
\EndIf

\If{$NN$ (nb of clients with high movement)$ \ge PE$}
\While{$NN \ge PE$}   \Comment{nb of high movement clients}
\State swap (high move $C$, not high move $C$)
\EndWhile

\EndIf

\EndWhile
\State\textbf{Return:} Repaired solution
\end{algorithmic}
\end{algorithm}

\section{EXPERIMENTS AND SIMULATION RESULTS}
In this section, we provide first a description of the data set used, the FL setup, the centralized model, the distribution of data over the clients, and the baseline approaches. Afterward, we state our results followed by a discussion and interpretation.

\subsection{Dataset}

The MOBILE DATA CHALLENGE (MDC) \cite{mario14} dataset is used in our learning. It consists of continuous records illustrating the movements of clients recorded over a period of time. We extracted from this dataset several features to help us predict the next place the clients are heading to. Global Positioning System (GPS) features were used to cluster 20 places and 6 area locations in a specific region on the map. Each area location contains different places. Time features were extracted and represented in several features like: day, month, year, weekend or weekday. We added an important feature to track the duration of a visit to a place by those users by calculating The duration of stay. Furthermore, frequency and rate of visits are extracted to track the locations of presence every month. Finally, the label class is added, corresponding to the next place a client visited in a specific duration of time.  

\subsection{Federated Learning Setup}
In order to run FL processes, we used Localfed \cite{marioo15}, which is a FL framework based on the FedML framework \cite{fedmll}. Localfed is known for its feasibility in applying FL applications using built-in components. These components are implemented in Python. It requires different parameters to control the behavior of the model. The trainer manager is responsible to define how the trainers are running, along with passing some parameters to define the number of epochs and loss. The aggregator instance is responsible to merge the collected models into a global model. The metrics component is responsible to test the model accuracy and loss on testing data after each round. Moreover, Trainers-data-dict defines the data of each client saved in dictionaries. After that, we define some parameters like the number of rounds, desired accuracy, train, and test data ratios.

Other components were added by us like Data Distribution and Base Model. The Data Distribution component was added to efficiently distribute the data to the clients. Since we are dealing with real-life dataset and specifically user-based. We assigned to each client the data of one user available in the dataset. This was done to simulate real-life scenario in which clients/users have the data on their side. The Base Model component is added in order to run any machine learning (ML) model under FL. 

\subsection{Centralized Model}
Before having a global model to be shared with the clients, we built a centralized model where the full training data has been used together. After many rounds of tuning hyper-parameters, Deep Neural Network \cite{neuraln} is used with three hidden layers of size 128 and 256. "Relu" and "softmax" activation functions are used. The input layer size is equal to the number of features we have and the output layer size is equal to the number of places in the dataset. The Adam optimizer is used with “categorical-cross entropy” loss function.

\subsection{Distribution of Data Over Clients}
There are 100 users in the dataset. Each client in our FL model is assigned the data of one user in the dataset. The data records fall between 200 and 1500 records. Since the problem is a multi-class classification task, the target feature defines the next place a user is going to by placing the Id number of this place. This partition follows a Non-iid data distribution since clients have different data partitions and class labels. 
In each learning round, the portion of deployed clients $C$ is decided by the optimization model. The set $C$ is not static since having less number of clients at early stages and more later is better for the learning process \cite{mario5}. This concept is implemented in our optimization model, so the number of deployed clients in each FL round is decided by the "Client Deployment" component represented in section IV.
\begin{figure}[]
	\centering
	\includegraphics[width=0.5\textwidth]{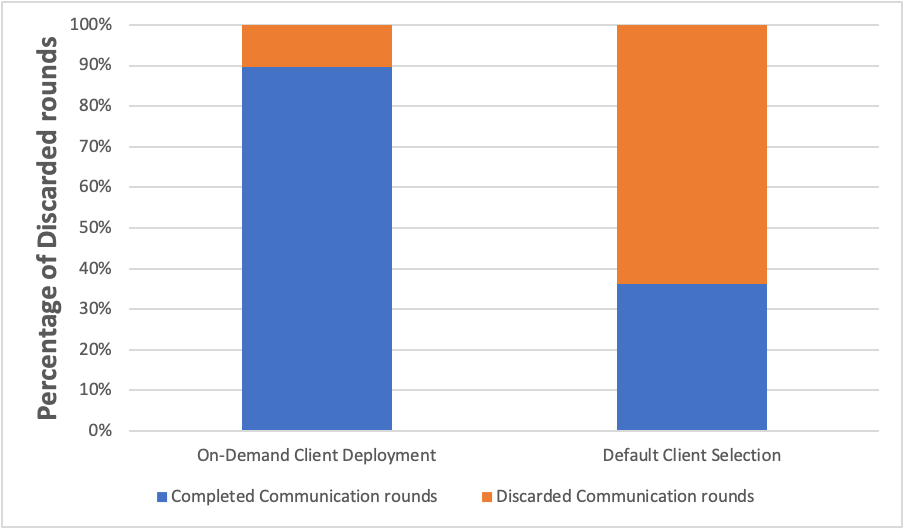}
	\caption{Completed vs discarded rounds in 50 FL rounds}
	\label{roundss}
\end{figure}

\subsection{Experiments and results}
We compared our approach to some of previous work:

\begin{enumerate}
    \item The static environment where constrained devices are (1) available in pre-defined locations and (2) not available to serve \cite{mario1}.
    \item The original FL selection problem (VanillaFL) \cite{mario12}, where clients participating in each round are randomly chosen.
    \item The centralized global model \cite{neuraln}.
\end{enumerate}

To prove the efficiency of our architecture, we conduct multiple scenarios answering multiple questions: 
\begin{enumerate}
    \item How does the availability of learning devices affect the learning process.
    \item How does the on-demand client deployment help the process of learning?
    \item How many rounds did the global model take to converge?
    \item How many devices are able to participate and finish their tasks without dropping out from the learning round?
    \item Is the network very congested?
    \item How does diversity and volume of data help the learning process while having a Non-iid dataset?
\end{enumerate}
In order to conduct the analysis and comparison for all the experiments, we applied the scenarios five times and took their averages into reference.
\begin{figure}[]
	\centering
	\includegraphics[width=0.5\textwidth]{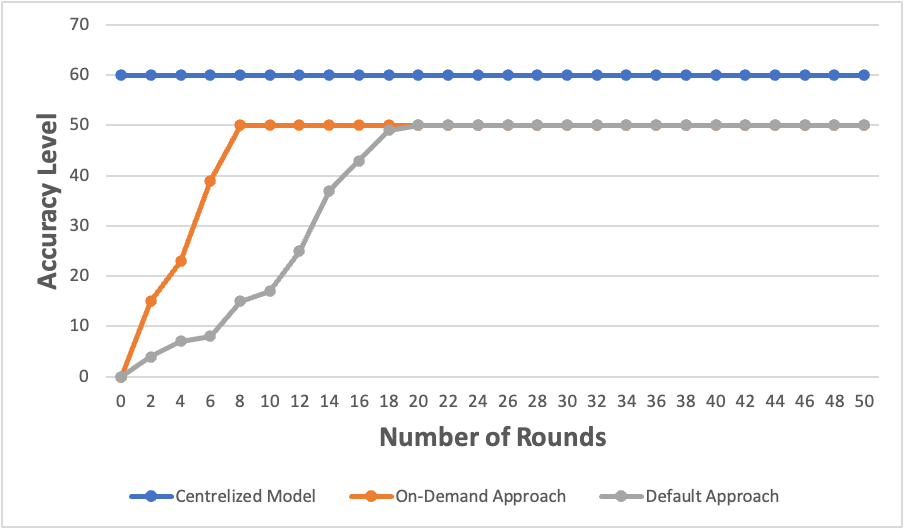}
	\caption{Test Accuracy with respect to the number of rounds}
	\label{m1}
\end{figure}

\begin{enumerate}

 \item We start with an environment rich in static and dynamic devices. In such places, devices are generating a huge volume of data that can be used for learning. However, most of these devices can be busy or do not have enough capabilities to be used as clients in FL. This results in postponing the process of learning or even neglecting the chance of having a ML process applied to them. On the other hand, as shown in Figure \ref{nbofclients}, using containerization technology to deploy docker containers on these machines results in having more available devices in the area. The containerization technology was proved to be efficient and more lightweight \cite{mario1} than virtual machines. For that, our architecture can make use of any constrained devices in this environment to serve as a client in each FL round. We start by having zero available clients, and as can be seen that by deploying containers on the volunteering devices more devices are becoming available. It is also noticed that the use of dynamic devices moving inside and outside the environment (area). Such devices can be used if reachable throughout the learning round. This process is available in our architecture since the "Client Deployment" component deploys those dynamic clients if their average staying period in this area is greater than the threshold needed to finish a learning round. Hence, this allows more diversity of clients to choose from. Moreover, as shown in Figure \ref{nbofclients}, our architecture can make use of weak static and dynamic devices in each area to make them available for learning. Therefore, we can efficiently and on-demand deploy clients in any environment having volunteering devices.

\item Afterward, a scenario is taken where there is a need to predict the next place users are heading to in particular locations in order to advertise some products in the destination place. As shown in Figure \ref{m1}, we can see a decrease in the number of FL rounds compared to the random selection architecture. Since client devices participate in a small number of rounds throughout the day \cite{mario12}, our approach succeeds to have fewer rounds to reach convergence while deploying clients at the right time and location while taking into consideration multiple constraints. In Figure \ref{m1} the blue line shows an accuracy of 60\% which is the centralized model to represent the desired accuracy. We calculate the number of rounds our approach takes to converge versus the default random client selection. To converge, our approach took on average 6-8 learning rounds, while this number was increased to 14-18 when using the default method, taking into assumption that enough client devices exist. This clearly shows the importance of choosing quality-based clients and applying our objective functions.

In addition, it was reported in \cite{mario10} that resource utilization increases with the size of the dataset. In FL, it is essential to take resources consumption and the movement of clients into consideration to make less number of client devices drop out from the learning due to resource overwhelming or reachability issues (for dynamic clients). Based on the available devices in an area, our architecture tries to deploy a set of these clients to participate in each round. As previously explained, by using our architecture there is a higher number of available clients. Therefore, there are more clients to choose from along with higher diversity. The number of deployed clients by the proposed model starts with a small set of clients which is 5 and increases this number in each round to reach 15-20.
The VanillaFL framework tries to take $[K \times C]$ set of clients. So a constant portion of clients is selected even though they might not be ready or prepared for the learning. On the other hand, our proposed architecture filters the clients and deploys the best set of clients to participate in each round if we have available clients, otherwise, our architecture has the ability to use any other constrained device to serve as a client.

\item Our approach succeeds to avoid any violation of the previously mentioned constraints so any deployed device has the ability to finish a learning round. On the other hand, using the default selection method, some of the selected clients did not respond back to the server. This depends on the constraints that our model takes into consideration and VanillaFL does not. Having a large number of clients failing to report back their updates yields the server to drop this round, which makes the overall process takes longer. Moreover, the constant number of selected clients by the FLVanilla framework leads to have fewer available clients to participate in other applications. On the other hand, our architecture makes use of a small amount of clients and makes the rest of the clients available for other ML applications to be applied in these areas. Our proposed model results in a few discarded rounds out of the 50 rounds, compared to the default method where the majority of the clients are misselected.

\item Besides, FL is famous with some network congestion, where a huge volume of exchanging messages and parameters must be done. Figure \ref{m1} shows that selecting fewer clients per round reduces the total number of rounds and the neglected rounds. Henceforth, our approach achieves less communication rounds. Moreover, using our orchestrators to exchange and deal with the deployed clients make the server less busy and congested to run other models and applications at the same time. As shown in Figure \ref{roundss}  we can see that the number of used communication rounds on the network using our proposed model is greater than of VanillaFL. The 10\% of the discarded rounds while using our proposed model results from: (1) if the client receives an invitation to deploy a service having higher priority than the ML service, or (2) if the deployed client changes its destination or its usual staying period in this place and area location. On the other hand, we can see that more than 50\% of the communication rounds are discarded while using VanillaFL framework, due to the random selection and the constant number of clients that the framework should choose even when we do not have such available or capable clients in the area. Furthermore, the default static devices can be far from the server and result in a long time to deliver the updates. Having close client devices supports in lowering the number of discarded communication rounds.

\begin{figure}[]
	\centering
	\includegraphics[width=0.5\textwidth]{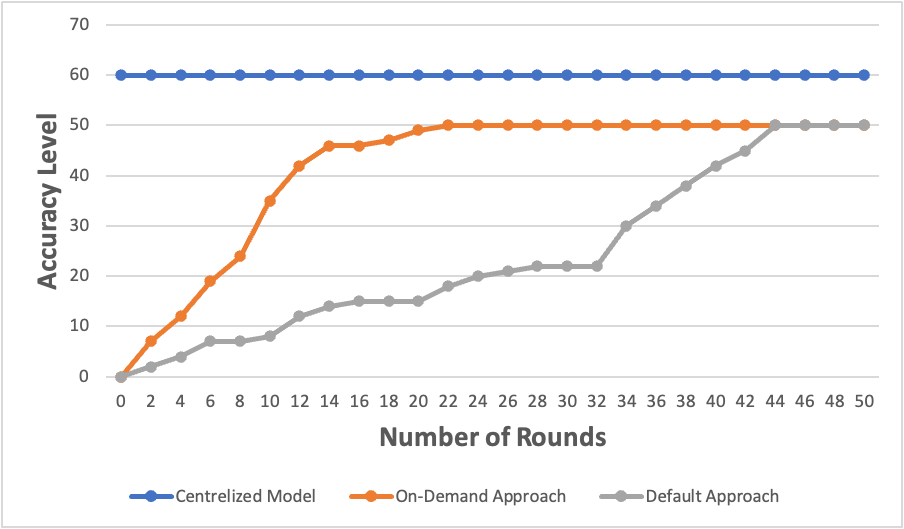}
	\caption{The evolution of the testing accuracy in respect to the number of available clients in Figure \ref{m3}}
	\label{m2}
\end{figure}
\begin{figure}[]
	\centering
	\includegraphics[width=0.5\textwidth]{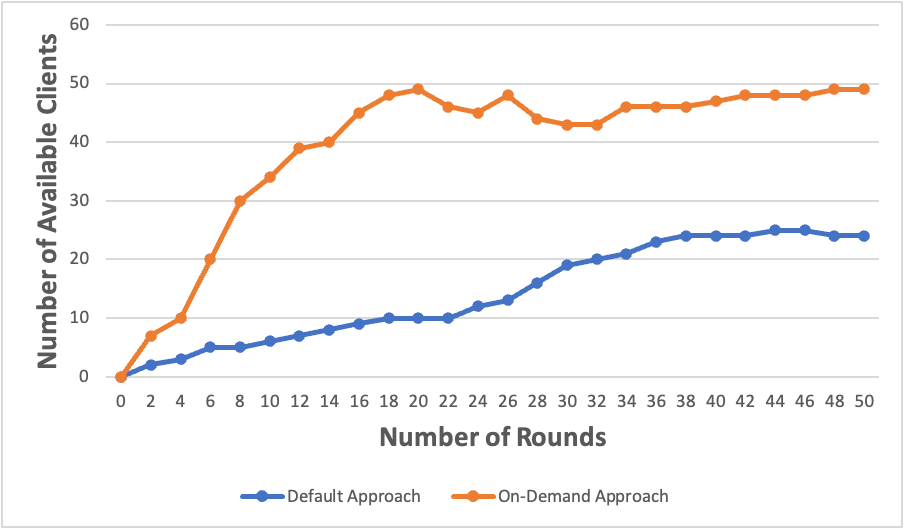}
	\caption{Number of available clients per round}
	\label{m3}
\end{figure}

Figure \ref{m2} and \ref{m3} show a comparison between the proposed architecture and the default one. Figure \ref{m2} shows the evolution of the accuracy level with respect to the number of clients present in Figure \ref{m3}. Following the analysis of figures, it can be concluded that a larger set of available clients, yield the model to converge faster compared to having few number of available devices.

\item In this part, we study the volume of data generated from our architecture and the diversity of clients. 
we take a scenario where having high movements with no available clients to participate in the learning. We apply our architecture to deploy clients on-the-fly and make use of the data generated. We can see in Figure \ref{fig4} the increase in the volume of data available for learning. This number is increasing while having the ability to deploy volunteers to serve as clients. The orchestrators play an important role while monitoring the average movements and visited places by the users. Once a high movement of devices is noticed in their area, a request is sent to the server to favor this area for some client deployments. Having high activities of devices in a location leads to generating a large volume of data that can be used in the learning. In addition, having a heterogeneity of clients results in having more diversity in the output class that we are trying to predict at each end. This helps the model to converge faster when having Non-iid data. As shown in Figure \ref{fig5}, choosing clients from different areas results in having more class labels, which makes the model learns faster.

\item Our priority system proposed in Section IV plays a great role in pushing the accuracy higher at early stages. Checking the quality of data a user has (without having access to the data) can be done by checking the client's contribution to the global model. If the client is not selected and has some good local accuracy compared to other clients, the type of data this client has can be concluded. Higher local accuracy means its testing data are somehow close to the data that the model trained on. Therefore, deploying such clients at early stages results in having more related training data to the model, which will push the accuracy at early stages as illustrated in Figure \ref{m1}. In later rounds, a priority algorithm running in the background helps to prioritize the clients fairly.

\item Lastly, using FL supports in preserving the privacy of clients participating in learning, where a 60\% accuracy is achieved in the centralized model and 50\% in the FL. The difference between the Federated and the Centralized models is logical since the privacy and data are not all in one place.  
\end{enumerate}

\section{Conclusion}
The primary issues addressed in this article are client availability and on-demand client deployment capability. We solve these issues by making use of the volunteering devices that are available everywhere in the designated regions. We utilize Kubeadm and Docker to manage on-the-fly deployments and to set up clients everywhere we go, whenever we need them. The server, the orchestrator, and the user are the three layers that make up our framework. We propose a multi-objective optimization solution to solve the client deployment problem. A heuristic
model based on Genetic Algorithm (GA) is elaborated for solving the multi-objective client/model deployment.
The experimental findings show the viability, efficiency, and improvement of applying the on-demand client deployment, where less time is required to achieve the desired precision compared with the default selection framework and the centralized model. Real-world scenarios are carried out to demonstrate the viability of client deployment in static regions and the advantages of having additional clients available in a learning environment. As future work, we plan to extend our architecture and optimize the client deployment by having multiple FL applications running in our environment.


%

\ifCLASSOPTIONcaptionsoff
  \newpage
\fi

\renewcommand*{\bibfont}{\small}

\printbibliography
\clearpage

\begin{IEEEbiography}[{\includegraphics[width=1in,height=1.25in,clip]{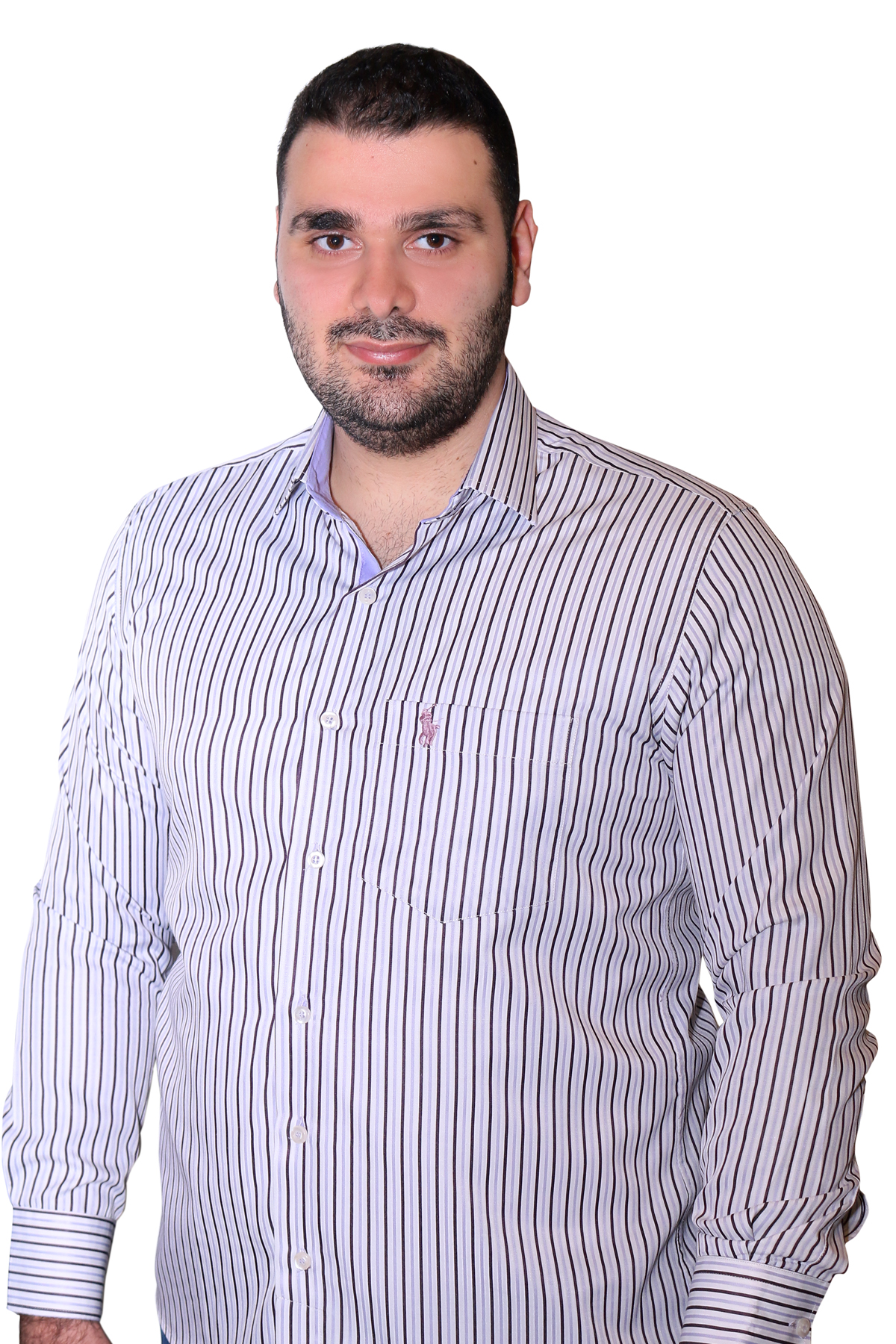}}]%
{\textbf{\textit{\textbf{Mario Chahoud}}}}
received his M.Sc. degree and B.S. degree in computer science from the Lebanese American University (LAU) based in Beirut, Lebanon. He worked as Research and Teaching assistant at the Lebanese American University. His research interests include fog and cloud computing, ML, and federated learning.
\end{IEEEbiography}
\vskip 0pt plus -1fil

\begin{IEEEbiography}[{\includegraphics[width=1in,height=1.25in,clip,keepaspectratio]{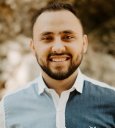}}]%
{\textbf{\textit{\textbf{Hani Sami}}}}
is currently a Ph.D. candidate at Concordia University, Institute for information Systems Engineering (CIISE). He received his M.Sc. degree in Computer Science from the American University of Beirut and completed his B.S. at the Lebanese American University. The topics of his research are fog computing, vehicular fog computing, and reinforcement learning. He is a reviewer of several prestigious conferences and journals.
\end{IEEEbiography}
\vskip 0pt plus -1fil

\begin{IEEEbiography}[{\includegraphics[width=1in,height=1.25in,clip,keepaspectratio]{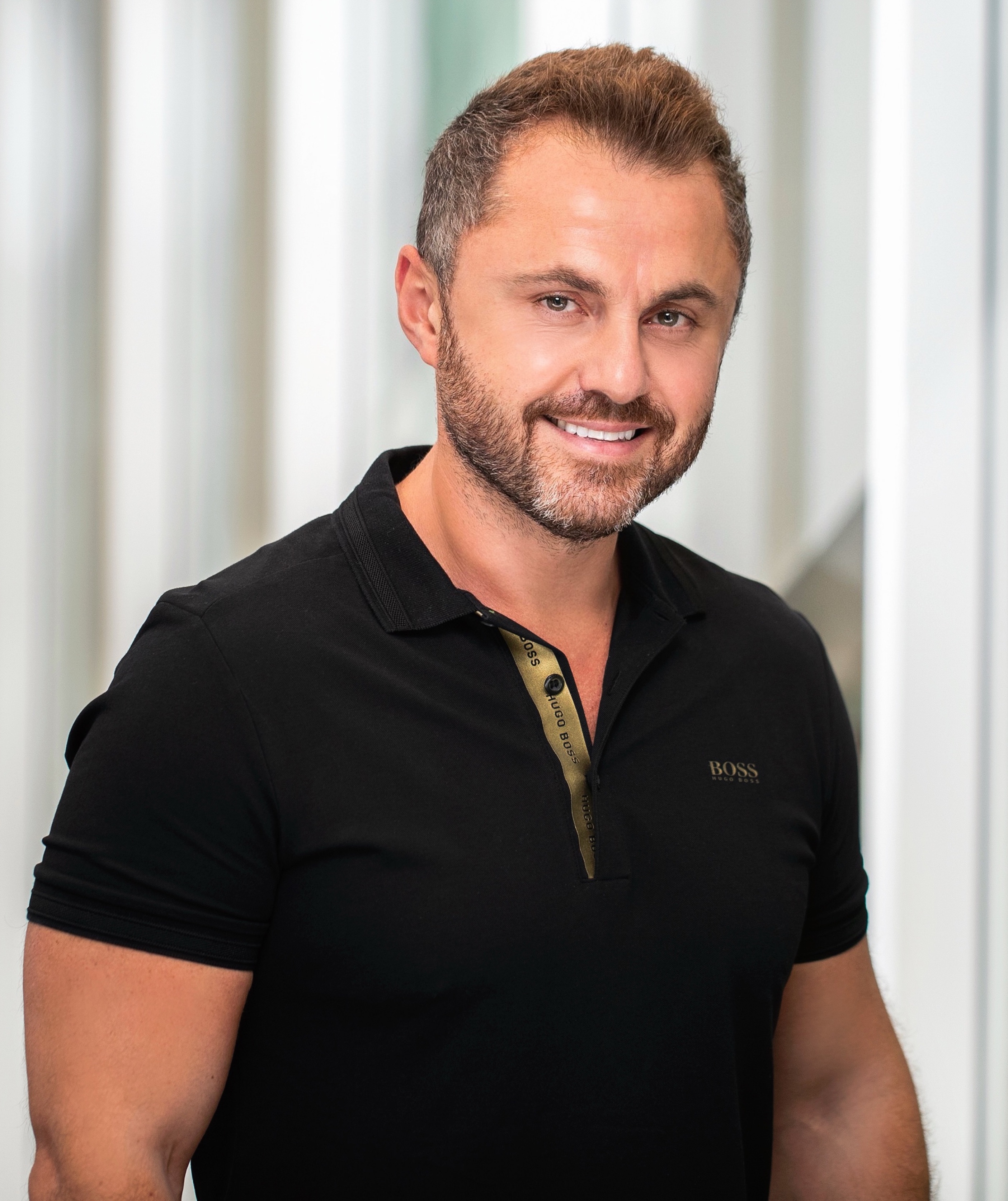}}]%
{\textbf{\textit{\textbf{Azzam Mourad}}}} received his M.Sc. in CS from Laval University, Canada (2003) and Ph.D. in ECE from Concordia University, Canada (2008). He is currently Professor of Computer Science and Founding Director of the Cyber Security Systems and Applied AI Research Center with the Lebanese American University, Visiting Professor of Computer Science with New York University Abu Dhabi and Affiliate Professor with the Software Engineering and IT Department, Ecole de Technologie Superieure (ETS), Montreal, Canada. His research interests include Cyber Security, Federated Machine Learning, Network and Service Optimization and Management targeting IoT and IoV, Cloud/Fog/Edge Computing, and Vehicular and Mobile Networks. He has served/serves as an associate editor for IEEE Transactions on Services Computing, IEEE Transactions on Network and Service Management, IEEE Network, IEEE Open Journal of the Communications Society, IET Quantum Communication, and IEEE Communications Letters, the General Chair of IWCMC2020, the General Co-Chair of WiMob2016, and the Track Chair, a TPC member, and a reviewer for several prestigious journals and conferences. He is an IEEE senior member.
\end{IEEEbiography}

\vskip 0pt plus -1fil

\begin{IEEEbiography}[{\includegraphics[width=1in,height=1.25in,clip,keepaspectratio]{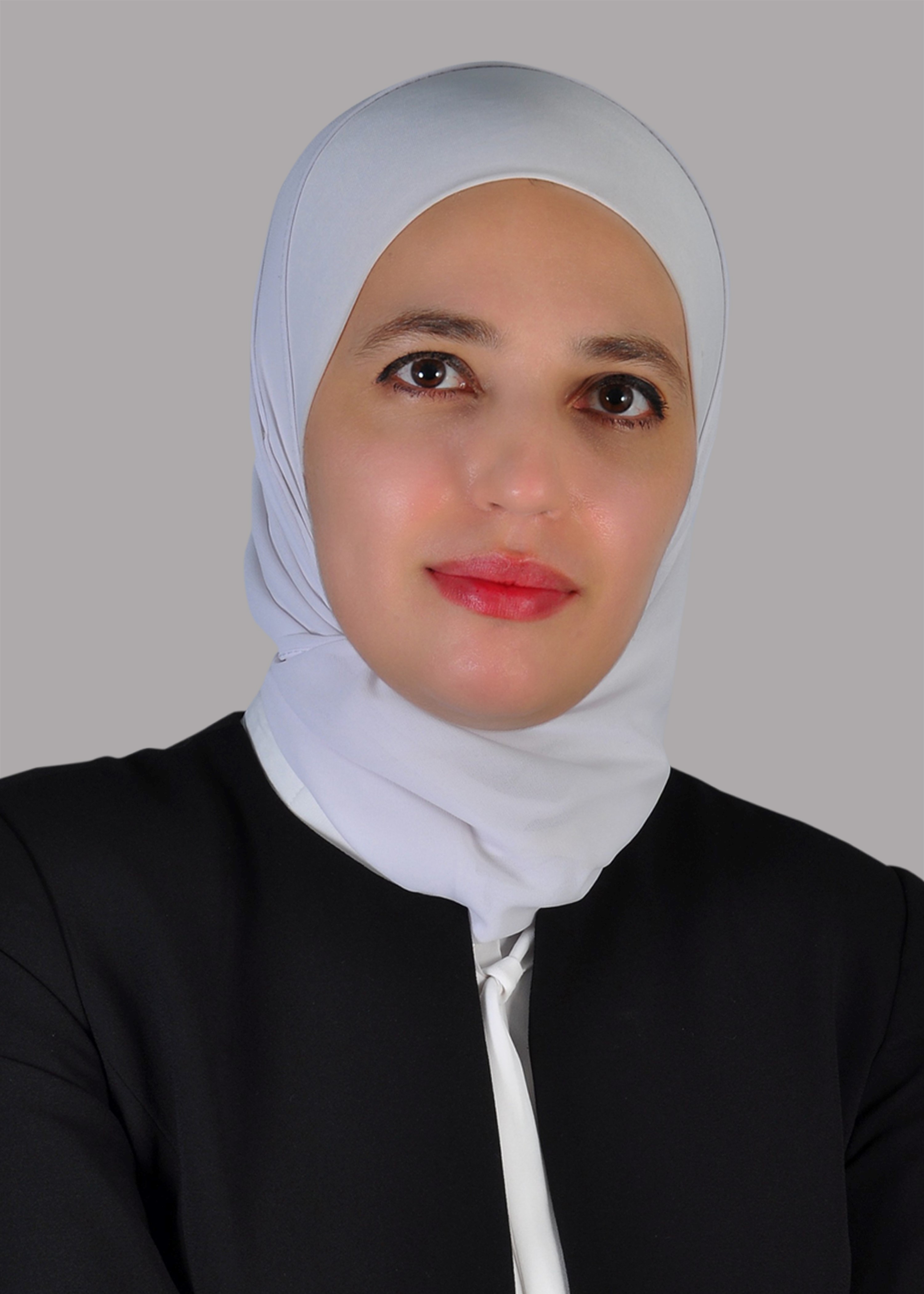}}]%
{\textbf{\textit{\textbf{Safa Otoum}}}}
(M’19) is an assistant professor of computer engineering in the College of Technological Innovation (CTI), Zayed University, United Arab Emirates. She received her M.A.Sc. and Ph.D. degrees in computer engineering from the University of Ottawa, Canada, in 2015 and 2019, respectively. Her research interests include blockchain applications, applications of ML and AI, IoT, and intrusion detection and prevention systems. She is a registered Professional Engineer (P.Eng.) in Ontario.
\end{IEEEbiography}
\vskip 0pt plus -1fil

\begin{IEEEbiography}[{\includegraphics[width=1in,height=1.25in,clip,keepaspectratio]{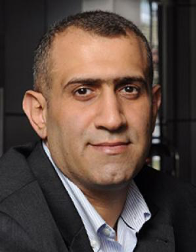}}]%
{\textbf{\textit{\textbf{Hadi Otrok}}}}
(senior member, IEEE) received his Ph.D. in ECE from Concordia University. He holds a Full professor position in the department of Electrical Engineering and Computer Science (EECS) at Khalifa University, an affiliate associate professor in the Concordia Institute for Information Systems Engineering at Concordia University, Montreal, Canada, and an affiliate associate professor in the electrical department at Ecole de Technologie Superieure (ETS), Montreal, Canada. He is an associate editor at: IEEE TNSM, Ad-hoc networks (Elsevier), and IEEE TSC. His research interests include: Blockchain, reinforcement learning, Federated Learning, crowd sensing and sourcing, ad hoc networks, and cloud and fog security.
\end{IEEEbiography}
\vskip 0pt plus -1fil
\begin{IEEEbiography}[{\includegraphics[width=1in,height=1.25in,clip,keepaspectratio]{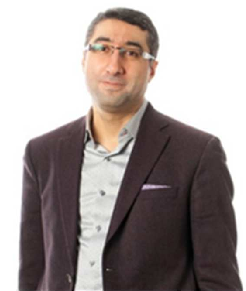}}]%
{\textbf{\textit{\textbf{Jamal Bentahar}}}}
received the Ph.D. degree in computer science and software engineering from Laval University, Canada, in 2005. He is a Professor with Concordia Institute for Information Systems Engineering, Concordia University, Canada. From 2005 to 2006, he was a Postdoctoral Fellow with Laval University, and then NSERC Postdoctoral Fellow at Simon Fraser University, Canada. He was an NSERC Co-Chair for Discovery Grant for Computer Science (2016-2018). He is a visiting professor at Khalifa University of Science and Technology. His research interests include the areas of computational logics, reinforcement learning, multi-agent systems, service computing, game theory, and software engineering.
\end{IEEEbiography}

\vskip 0pt plus -1fil
\begin{IEEEbiography}[{\includegraphics[width=1in,height=1.25in,clip,keepaspectratio]{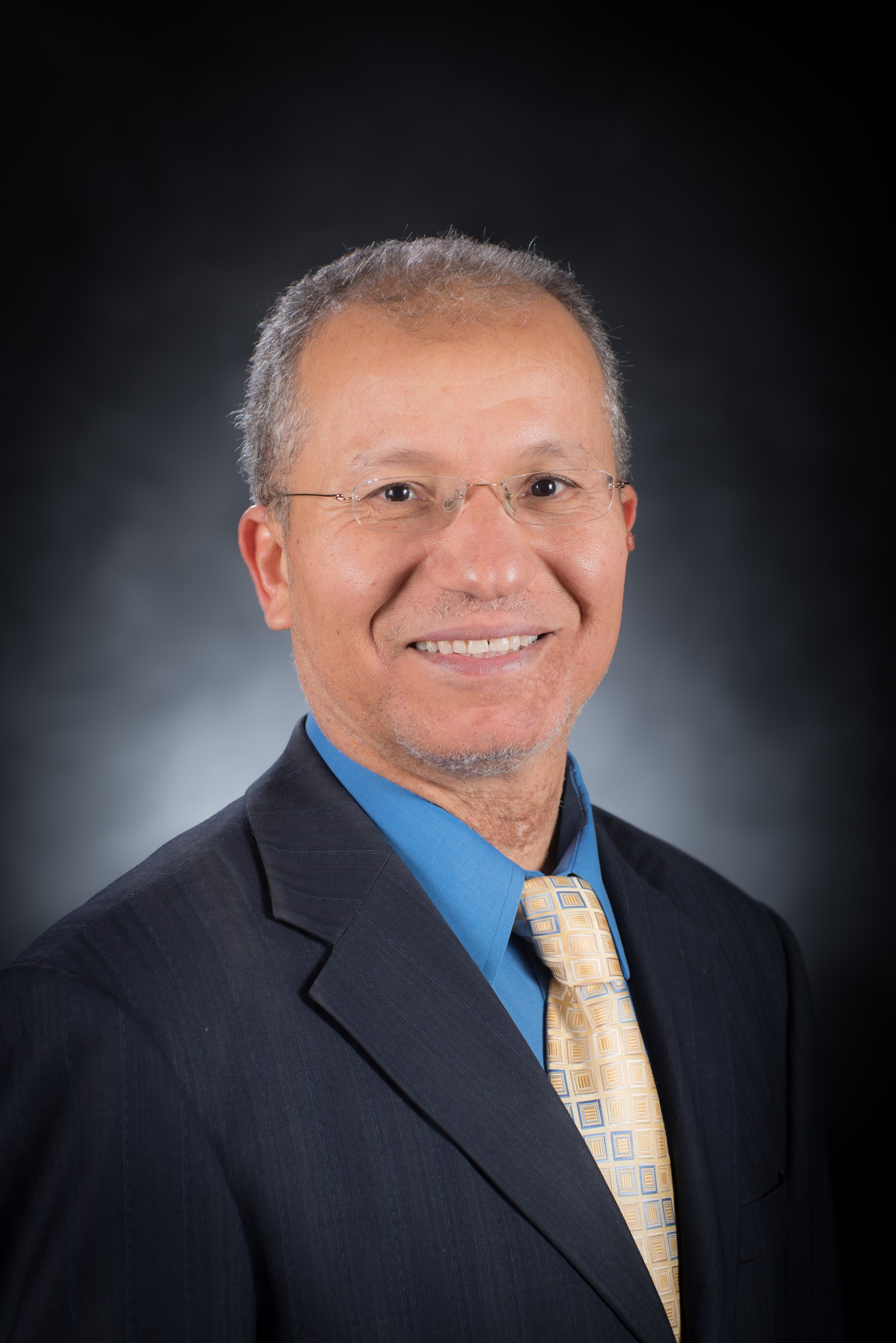}}]%
{\textbf{\textit{\textbf{Mohsen Guizani}}}}
(Fellow, IEEE) received the BS (with distinction), MS and PhD degrees in Electrical and Computer engineering from Syracuse University, Syracuse, NY, USA in 1985, 1987 and 1990, respectively. He is currently a Professor of Machine Learning and the Associate Provost at Mohamed Bin Zayed University of Artificial Intelligence (MBZUAI), Abu Dhabi, UAE. Previously, he worked in different institutions in the USA. His research interests include applied machine learning and artificial intelligence, Internet of Things (IoT), intelligent autonomous systems, smart city, and cybersecurity. He was elevated to the IEEE Fellow in 2009 and was listed as a Clarivate Analytics Highly Cited Researcher in Computer Science in 2019, 2020 and 2021. Dr. Guizani has won several research awards including the “2015 IEEE Communications Society Best Survey Paper Award”, the Best ComSoc Journal Paper Award in 2021 as well five Best Paper Awards from ICC and Globecom Conferences. He is the author of ten books and more than 800 publications. He is also the recipient of the 2017 IEEE Communications Society Wireless Technical Committee (WTC) Recognition Award, the 2018 AdHoc Technical Committee Recognition Award, and the 2019 IEEE Communications and Information Security Technical Recognition (CISTC) Award. He served as the Editor-in-Chief of IEEE Network and is currently serving on the Editorial Boards of many IEEE Transactions and Magazines. He was the Chair of the IEEE Communications Society Wireless Technical Committee and the Chair of the TAOS Technical Committee. He served as the IEEE Computer Society Distinguished Speaker and is currently the IEEE ComSoc Distinguished Lecturer. 
\end{IEEEbiography}

\end{document}